\documentclass[letterpaper, 10 pt, conference]{ieeeconf}  

\IEEEoverridecommandlockouts                              

\usepackage[table, dvipsnames]{xcolor}
\usepackage{amsmath} 
\usepackage{graphicx}
\usepackage{adjustbox}
\usepackage{multicol}
\usepackage{tabularx}
\usepackage{caption}
\usepackage{amssymb}
\usepackage{cellspace}
\usepackage{hyperref}
\usepackage{cite}
\usepackage{multirow}
\usepackage{float}
\usepackage{booktabs}
\usepackage{subcaption}
\usepackage{soul}
\let\labelindent\relax
\usepackage{enumitem}
\usepackage{xspace}

\definecolor{pastelDarkestCoral}{HTML}{FF704D}
\definecolor{pastelDarkestLavender}{HTML}{653be3}
\definecolor{pastelDarkestSkyBlue}{HTML}{ed2fb4}
\definecolor{pastelDarkestMint}{HTML}{2ca02c}

\usepackage{tikz}
\newcommand{\colunderline}[2]{%
  \tikz[baseline=(X.base)]{
    \node[inner sep=0pt, outer sep=0pt, anchor=base] (X) {\ensuremath{#2}};
    \draw[#1, line width=0.5pt] 
      ([yshift=-1.2pt]X.south west) -- ([yshift=-1.2pt]X.south east);
  }%
}
\newcommand{\textcolunderline}[2]{%
  \tikz[baseline=(X.base)]{
    \node[inner sep=0pt, outer sep=0pt, anchor=base, text=black] (X) {\textit{#2}};
    \draw[#1, line width=0.5pt] 
      ([yshift=-1pt]X.south west) -- ([yshift=-1pt]X.south east);
  }%
}

\newcommand{\coolname}{DAGDiff\xspace}

\title{\LARGE \bf
\coolname: Guiding Dual-Arm Grasp Diffusion to Stable and Collision-Free Grasps
}

\author{Md Faizal Karim$^{1*}$, Vignesh Vembar$^{1*}$, Keshab Patra$^{1}$, Gaurav Singh$^{2}$, K Madhava Krishna$^{1}$
\thanks{$^*$ Equal Contribution}
\thanks{$^{1}$ Robotics Research Center, IIIT Hyderabad}%
\thanks{$^{2}$ Brown University}%
}

\begin{document}

\maketitle
\thispagestyle{empty}
\pagestyle{empty}

\begin{abstract}

Reliable dual-arm grasping is essential for manipulating large and complex objects but remains a challenging problem due to stability, collision, and generalization requirements. Prior methods typically decompose the task into two independent grasp proposals, relying on region priors or heuristics that limit generalization and provide no principled guarantee of stability. We propose \coolname, an end-to-end framework that directly denoises to grasp pairs in the $SE(3) \times SE(3)$ space. Our key insight is that stability and collision can be enforced more effectively by guiding the diffusion process with classifier signals, rather than relying on explicit region detection or object priors. To this end, \coolname integrates geometry-, stability-, and collision-aware guidance terms that steer the generative process toward grasps that are physically valid and force-closure compliant. We comprehensively evaluate \coolname through analytical force-closure checks, collision analysis, and large-scale physics-based simulations, showing consistent improvements over previous work on these metrics. Finally, we demonstrate that our framework generates dual-arm grasps directly on real-world point clouds of previously unseen objects, which are executed on a heterogeneous dual-arm setup where two manipulators reliably grasp and lift them. Project Page: \href{https://dag-diff.github.io/dagdiff/}{dag-diff.github.io/dagdiff/}

\end{abstract}

\section{Introduction}

Manipulating large, dual-arm relevant objects such as monitors, boxes, or chairs requires not only feasible grasps, but also reasoning about force balance and stable interaction between both arms. Imagine the task of picking up a monitor. Humans instinctively place their hands on the opposite sides of the monitor instead of grasping it at random points to balance forces and torques, ensuring stability. For robots, however, acquiring this kind of coordination is far more complex \cite{dual_arm_manipulation}. Developing this sense of dual-arm stability is essential for moving beyond single-arm grasping toward coordinated, physically robust manipulation of real-world objects \cite{robot_learning_survey, trends_in_manipulation, robotic_grasping_classical, flat}.

While grasp pose generation has been explored extensively in the community, most efforts largely focus on single-arm settings. Most methods \cite{dexnet2.0, gpd, contact_graspnet, anygrasp, grasp_anything, 6dof_graspnet} follow a general recipe of curating a paired dataset consisting of ground truth grasps evaluated using physics simulators, followed by training encoder-decoder style models in a supervised setting. Recently, diffusion models have emerged as powerful generative frameworks for robotic grasping \cite{cgdf, se3diff, graspgen, dex_diffuser, dex_grasp_anything} due to their ability to model complex multimodal distributions. This enables them to sample diverse, high-quality grasp poses either uniformly across the object or constrained to specific parts.

While these methods have improved robustness and grasp quality on complex shapes, they are designed for single-arm grasps and lack mechanisms to ensure dual-arm stability. Moreover, extending these methods to dual-arm grasping is non-trivial, as exhaustive pair search is costly and naive single-arm extensions often yield unstable solutions \cite{dg16m}. Furthermore, these methods do not explicitly account for collisions and often produce grasps that intersect the object surface, a problem that becomes increasingly critical for larger and more geometrically complex shapes. A possible workaround would be to increase the resolution of the point cloud or latent representation to capture finer surface details, but this would greatly increase computation without guaranteeing collision-free grasps.
\definecolor{darkgreen}{HTML}{08a100}
\definecolor{darkred}{HTML}{ab0000}
\begin{figure}[t]
    \centering
    \captionsetup{font=footnotesize}
    \includegraphics[width=\linewidth]{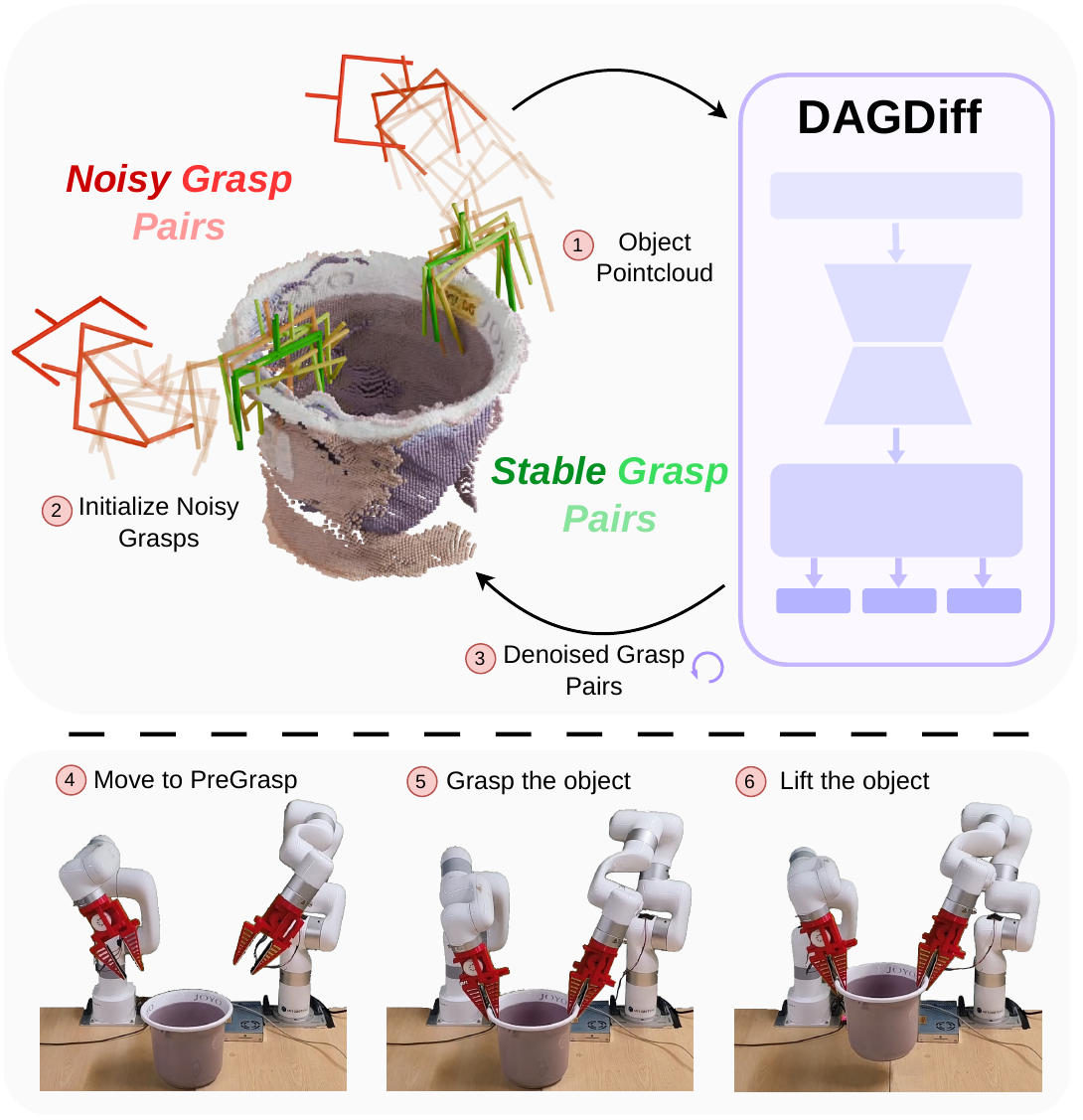}
    \caption{We introduce \coolname: \underline{D}ual-\underline{A}rm \underline{G}rasp \underline{Diff}usion, a diffusion framework in $SE(3) \times SE(3)$ that takes an object point cloud and denoises noisy grasp pairs (in shades of \textcolor{darkred}{red}) into stable, collision-free dual-arm grasps (in shades of \textcolor{darkgreen}{green}), guided by multi-head outputs. These predicted grasps are further validated through real-world dual-arm executions, where objects are grasped and lifted successfully. }
    \label{fig:teaser}
\end{figure}

In this work, we introduce \textbf{\coolname: \underline{D}ual-\underline{A}rm \underline{G}rasp \underline{Diff}usion}, an end-to-end dual-arm grasp generation framework that leverages diffusion models guided by classifier signals. Our method extends $SE(3)$ diffusion to the dual-arm setting to generate grasp pairs that are simultaneously stable under dual-arm force closure and collision free with respect to the object's surface.

We frame dual-arm grasp generation as the task of generating two grasps on the object point cloud, each falling in a suitable region, such that they are jointly stable and physically valid (by physically valid we mean grasps that are collision free, and make stable surface contact with the object). One of the key challenges is region selection: heuristic approaches such as choosing farthest regions \cite{cgdf} often fail when those regions itself are physically incompatible, while VLM-based reasoning \cite{unidiff} remains limited in 3D and physical understanding \cite{phys_bench, mech_reasoning}. It is further hindered by the fact that graspable regions rarely have semantic names, leaving no reliable basis for prediction. In contrast, our approach does not rely on region-specific training but instead learns suitable grasp regions implicitly from guidance signals, and we observe that it naturally discovers the right pairs of regions for stable dual-arm grasps (given in supplementary video).
Specifically, a \textbf{force-closure module} distinguishes stable from unstable grasp pairs and provides gradients that bias the diffusion process toward stability, while a \textbf{collision module} identifies grasp–object intersections and pushes generated grasps away from collisions. Together, these signals guide the diffusion model to diverse, stable, and physically valid dual-arm grasps. 

Our evaluation demonstrates the effectiveness of the proposed method in generating stable grasps within a dual-arm setup. Analytical evaluation based on dual-arm force-closure criteria \cite{dg16m} confirms that the generated grasp pairs satisfy fundamental stability requirements, while physics simulation-based evaluation \cite{isaac_gym} highlights the robustness of our approach across diverse objects and grasp configurations. Finally, real-world demonstrations show that our framework, trained entirely on synthetic data, transfers effectively to real point clouds, producing physically realizable dual-arm grasps on previously unseen objects like cooking utensils, buckets, drones etc as shown in Figure \ref{fig:teaser}. To summarize the contributions:
\begin{enumerate}
    \item We present \textbf{\coolname}, the first framework to the best of our knowledge, for dual-arm grasp generation that extends $SE(3)$ diffusion with guidance signals, enabling the synthesis of grasp pairs that are both force-closure stable and collision-free on large, geometrically complex objects.
    
    \item Unlike prior methods that rely on region identification using VLMs or geometric heuristics, our architecture employs geometry-, stability-, and collision-aware multi-head outputs that directly guide the diffusion process toward valid regions of the dual-arm grasp space (Figure \ref{fig:pipeline}). 

    \item We show substantial improvements over prior methods and adapted baselines through analytical metrics and large-scale simulations (Table \ref{tab:dual_arm_results}), and further validate reliable zero-shot transfer on a heterogeneous real-world dual-arm setup with real point clouds and previously unseen objects (Figure \ref{fig:teaser}). 
\end{enumerate}

\section{Related Works}

\subsection{Dual-Arm Grasping and Stability}
Dual-arm grasping requires two parallel-jaw grasps that are not only individually stable but also jointly satisfy stability criteria such as force-closure, i.e., the contact forces must counteract any external wrench on the object \cite{intro_to_robotics}. Mesh-based approaches like \cite{da2, dg16m} address this by densely sampling single-arm grasps on object meshes and evaluatin g all grasp pairs with a dual-arm force-closure test. While this ensures stability in simulation, the reliance on complete meshes and exhaustive pair evaluation makes these methods impractical for real-world perception and deployment. To reduce this combinatorial complexity, CGDF \cite{cgdf} employs a part-guided diffusion strategy that generates grasps in the two farthest regions of the point cloud, forming dual-arm pairs. Further, UniDiffGrasp \cite{unidiff} extends this idea by incorporating a VLM to identify object parts for dual-arm grasping. However, these methods still treat the problem as combining two single-arm proposals without explicitly enforcing joint stability. In contrast, DualAfford \cite{dualafford} directly predicts dual-arm grasp poses through collaborative affordance learning, generating one gripper’s grasp conditioned on the other. While this captures inter-gripper dependencies, the method relies on object category-specific training and an intricate pipeline, which limits its generalization. To overcome these issues, we propose an end-to-end diffusion framework to implicitly learn the distribution of stable dual-arm grasps without relying on external region-proposals or object-centric pipelines.

\subsection{Diffusion Models for Grasp Generation}
Diffusion models, which are particularly suited for capturing multimodal distributions, have emerged as a powerful alternative to previous classical as well as deep-learning based methods for grasp synthesis \cite{graspit, intro_to_robotics, gpd, contact_graspnet, anygrasp, 6dof_graspnet}. SE3Diff \cite{se3diff} introduced diffusion in the $SE(3)$ space for sampling diverse single-arm grasps, and \cite{grasp_diffusion_network} implemented this idea for partial point clouds along with refinement using collision spheres. CGDF \cite{cgdf} extended diffusion in $SE(3)$ with improved feature representations for constrained grasping on complex shapes. More recently, \cite{graspgen} combined diffusion with transformers to scale grasp generation to large datasets with strong sim-to-real performance. Beyond single-arm settings, diffusion has also been explored for dexterous and multi-fingered hands, with recent works \cite{dex_diffuser, dex_grasp_diffusion, dex_grasp_anything} demonstrating its effectiveness for generating stable and generalizable grasps in high-DOF gripper settings. Together, these approaches demonstrate the flexibility of diffusion for both simple and high-DOF grasp generation. However, they
do not naturally extend to the dual-arm setting, where the
challenge is not only generating individually valid grasps but also ensuring their joint stability. In this work, we address this challenge by introducing guidance signals that steer the diffusion process toward grasp pairs that are both feasible and dual-arm stable.
\begin{figure*}[!t]
    \centering
    \vspace{3pt}
    \captionsetup{font=footnotesize}
    \includegraphics[width=\linewidth]{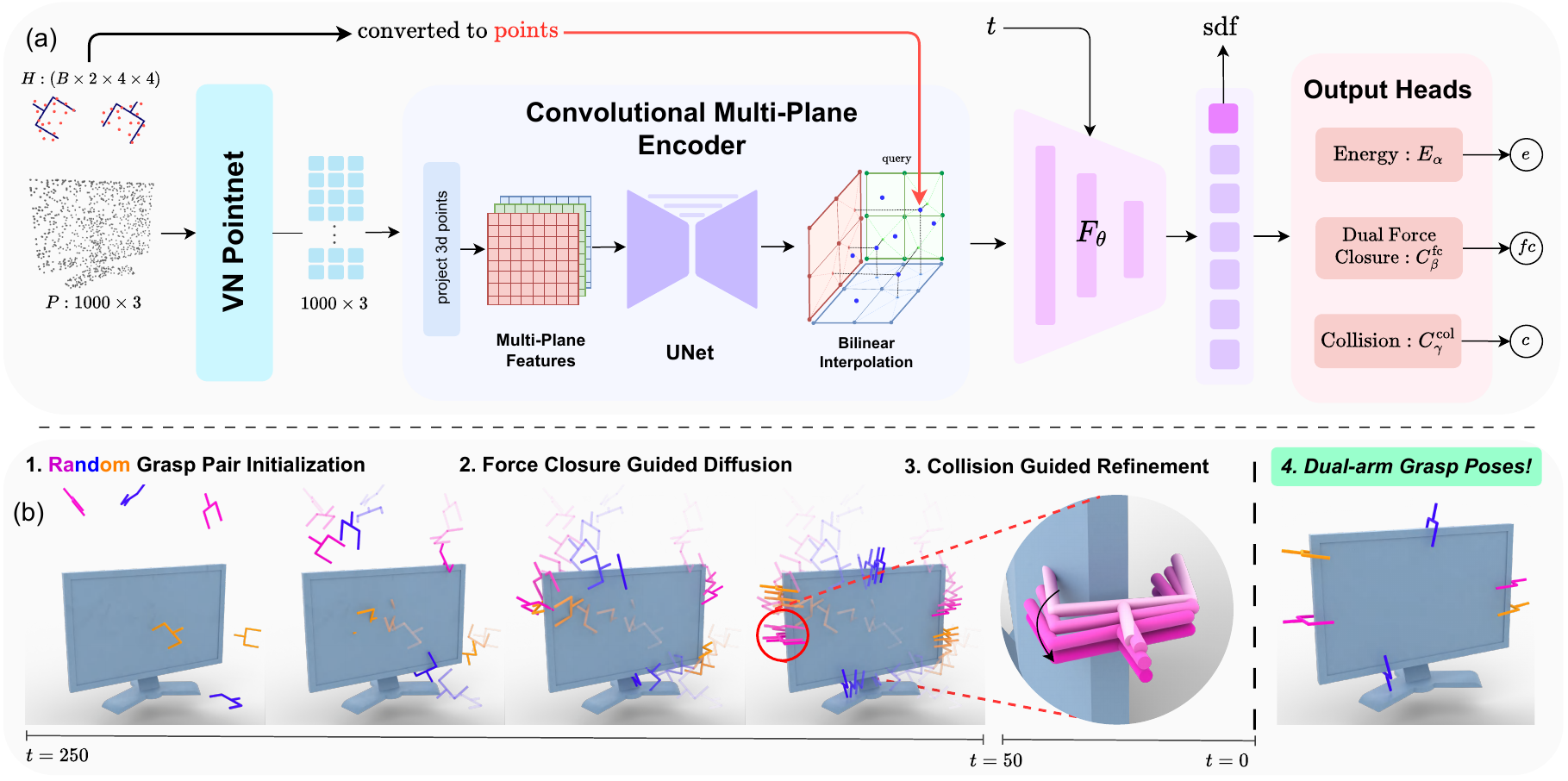}
    \caption{\textbf{Overview of the proposed method:} \textbf{(a)} Given an object point cloud $P$, our network encodes geometric features into dense feature maps. Next, randomly initialized dual-arm grasps $H$ are used to transform a fixed query cloud into query points, followed by feature sampling through bilinear interpolation. Conditioned on the noise step $t$, these features are passed through $F_\theta$, which predicts both the SDF of the query points and a feature vector. This vector is used by three output heads that predict energy $(E_\alpha)$, force-closure probability $(C_\beta^{\text{fc}})$, and collision probability $(C_\gamma^{\text{col}})$, jointly guiding the diffusion process. \textbf{(b)} At inference, denoising proceeds from random initializations ($t=250$) to refined grasps ($t=0$). The energy head drives the generative dynamics, while the force-closure and collision heads bias the generation until stable, collision-free dual-arm grasps emerge.}
    \label{fig:pipeline}
\end{figure*}

\section{Methods}
\label{sec:methods}
Given an object point cloud $P \in \mathbb{R}^{n\times3}$, our goal is to generate $M$ pairs of force closure stable and collision-free parallel-jaw grasp poses $H_i = (H_{i,1},H_{i,2})\in SE(3) \times SE(3), i \in [1, M]$ on $P$. The feasibility of $H$ is determined jointly by (i) whether each grasp maintains a stable, non-colliding contact with the object surface, and (ii) whether the pair jointly satisfies the dual-arm force-closure constraints \cite{dg16m}.
To address this, we formulate dual-arm grasp generation as a diffusion process in $SE(3) \times SE(3)$. Starting from random pairs of initial poses, we iteratively refine grasp candidates toward physically valid configurations using the score function learned by an energy-based model. In addition to it, we jointly train two classifier-guidance modules --- a force-closure classifier and a collision classifier, that provide gradient signals during inference. Together, these components enable the model to generate low-energy, dual-arm stable, and collision-free grasp pairs as shown in Figure \ref{fig:pipeline}.
To formalize diffusion in the $SE(3) \times SE(3)$ space, we introduce the following notation. The dual-arm logarithmic map $\operatorname{Logmap_2}\colon SE(3) \times SE(3) \to \mathbb{R}^{12}$ is defined as:
\begin{equation*}
    \textbf{v} = \operatorname{Logmap_2} \left(H\right) := \operatorname{Logmap}\left(H_1\right) \oplus \operatorname{Logmap}\left(H_2\right)
    \label{eq:logmap_notation}
\end{equation*} 
where $\oplus$ is the vertical concatenation operator. Similarly, we define the dual-arm exponential map $\operatorname{Expmap_2}\colon \mathbb{R}^{12} \to SE(3) \times SE(3)$ as:
\begin{equation*}
    H = \operatorname{Expmap_2} \left(\textbf{v}\right) := \left( \operatorname{Expmap}\left(\textbf{v}_{[:6]}\right), \operatorname{Expmap}\left(\textbf{v}_{[6:]}\right) \right)
    \label{eq:expmap_notation}
\end{equation*}
where $\textbf{v}_{[:6]}$ and $\textbf{v}_{[6:]}$ refer to the first and last six components of the vector $\textbf{v} \in \mathbb{R}^{12}$.

In this section, we first discuss the formulation of a diffusion-based dual-arm grasp generation model \textbf{(A)}, then formulate the classifier guidance modules for force-closure and collision \textbf{(B)} and finally outline the architecture of the complete generation model \textbf{(C)}.

\subsection{Diffusion-based Dual-arm Grasp Generation}
\label{sub_sec:dual_arm_diffusion}
We adapt the diffusion formulation of SE3Diff \cite{se3diff} to the dual-arm setting. A dual-arm grasp pose $H$ lies in $SE(3) \times SE(3)$, where each element of the pair lies in a Lie group that does not allow direct Euclidean operations. To address this, we map $H$ to a vector $\textbf{v} \in \mathbb{R}^{12}$  using $\operatorname{Logmap_2}$, which allows the diffusion process to operate in $\mathbb{R}^{12}$, while $\operatorname{Expmap_2}$ is used to convert perturbed samples back to $SE(3) \times SE(3)$. The denoiser is then defined as a vector field $s_\alpha$ that predicts a vector $d \in \mathbb{R}^{12}$ given a dual-arm grasp pose $H$, the object point cloud $P$, and the noise step $t \in [0,T)$, where $T$ is the total number of noise steps. 

\textbf{Energy Based Formulation}: In standard diffusion models, one option is to directly predict the added Gaussian noise at each noise step. However, learning the score function \cite{score_diffusion} has shown to yield more stable training and better likelihood modeling, since the score directly captures the structure of the noisy distribution. Formally, the score is defined as: $s(H, P, t) = \nabla_{H} \log p_t(H|P)$, where $p_t$ is the distribution of the noisy grasps at noise step $t$. 


Hence, we create an energy-based model $E_\alpha$, with learnable parameters $\alpha$, to learn a scalar energy landscape over 
\textcolunderline{pastelDarkestLavender}{dual-arm grasp} poses and an input object
\textcolunderline{pastelDarkestSkyBlue}{point cloud} along with the current
\textcolunderline{pastelDarkestMint}{noise step}. Formally, 
$E_\alpha\colon 
\colunderline{pastelDarkestLavender}{SE(3) \times SE(3)} \times 
\colunderline{pastelDarkestSkyBlue}{\mathbb{R}^{n\times3}} \times 
\colunderline{pastelDarkestMint}{\mathbb{R}} \rightarrow 
\mathbb{R}$.
The denoising vector field is then obtained as the negative gradient of the energy,  given by, 
\begin{equation}
    s_\alpha(H, P, t) = -\nabla_H E_\alpha(H, P, t) \in \mathbb{R}^{12}
\end{equation}
This formulation is equivalent to score-based diffusion, with the advantage that the energy function also provides a natural way to rank grasp candidates. Lower energy indicates grasp pairs that are physically grounded on the object rather than arbitrarily floating in free space.

\textbf{Forward and Reverse diffusion}: During training, the forward diffusion process perturbs the ground truth dual-arm grasp poses by adding noise in $\mathbb{R}^{12}$. The perturbed vector is converted back to $SE(3) \times SE(3)$ using the $\operatorname{Expmap_2}$ operation as:
\begin{equation}
\tilde{H}_t = \operatorname{Expmap_2}\!\left( \operatorname{Logmap_2}(H) + \epsilon_t \right), 
\label{eq:expmap-logmap}
\end{equation}
where $\epsilon_t \sim \mathcal{N}\!\left(0, \sigma_t^2 I_{12} \right)$ and $\sigma_t$ is the standard deviation at noise step $t$. This forward process progressively produces noisier versions of the grasp poses, which are used to train the model to denoise. At inference, the denoising process iteratively removes noise using a Langevin-style update. Given a noisy sample, $H_t$, the reverse step is defined as: 
\begin{equation}
H_{t-1} = \operatorname{Expmap_2}\!\left(
    \frac{\eta_t^2}{2}\, s_\alpha(H_t, P, t) 
    + \eta_t \epsilon
\right) H_t, 
\label{eq:expmap-step}
\end{equation}
where $\epsilon \sim \mathcal{N}(0, I_{12})$ and $\eta_t \geq 0$ is a step-dependent coefficient controlling the update magnitude. 

\textbf{Loss function:} The diffusion network is trained using the regular denoising loss function as the L1 norm between the normalized sampled noise $\frac{\epsilon_t}{\sigma_t}$ and predicted vector field $s_\alpha(H_t, P, t)$. Formally, 
\begin{equation}
\mathcal{L}_{\text{diff}} 
= \bigg\| s_\alpha(H_t, P, t) - \frac{\epsilon_t}{\sigma_t} \bigg\|_1
\label{eq:ldiff}
\end{equation}

\subsection{Classifier Guidance Modules}

Naively training a diffusion model in the dual-arm setting leads to poor generalization, since there is no explicit constraint enforcing the generation of stable dual-arm grasps. To address this, we adopt classifier-guided diffusion \cite{dhariwal_classifier_guidance}, which steers the generative process toward desired regions of the sample space by incorporating classifier log-likelihood gradients at each reverse step, thereby biasing the generation toward samples with specific properties.

We employ two classifier-guidance modules. The first is a \textbf{force-closure classifier} $C_{\beta}^{\text{FC}} = p(y=1|H, P; \beta)$ with learnable parameters $\beta$, which predicts the probability that a dual-arm grasp pair $H$ on object point cloud $P$ satisfies the force-closure stability criterion. During inference, its gradient with respect to the grasp pose guides the diffusion process toward force-closure stable configurations. The second is a \textbf{collision classifier} $C_{\gamma}^{\text{Col}} = p(y=1|H, P ;\gamma)$ with learnable parameters $\gamma$, which predicts the probability that a grasp pose $H$ is in collision with the object point cloud $P$. During inference, its gradient is used to refine generated candidates by pushing them away from collisions with the object. 


Both classifiers are trained with the standard binary cross-entropy loss:
\begin{align}
\mathcal{L}_{\text{fc}}  &= \operatorname{BCE}\!\left(C_{\beta}^{\text{FC}}(H, P), \, y_{\text{fc}}\right) \label{eq:lfc} \\
\mathcal{L}_{\text{col}} &= \operatorname{BCE}\!\left(C_{\gamma}^{\text{Col}}(H, P), \, y_{\text{col}}\right) \label{eq:lcol}
\end{align}
where $y_{\text{fc}} \in \{0, 1\}$ indicates whether the grasp pair satisfies force closure, and 
$y_{\text{col}} \in \{0, 1\} \times \{0, 1\}$ denotes which grasp(s) in the pair collide with the object. 
The overall training objective then combines the diffusion loss (Equation \ref{eq:ldiff}) with the classifier losses (Equation \ref{eq:lfc}, \ref{eq:lcol}) as:
\begin{equation}
\mathcal{L} = \mathcal{L}_{\text{diff}}  + \mathcal{L}_{\text{fc}} + \mathcal{L}_{\text{col}}.
\label{eq:total-loss}
\end{equation}

At inference time, the gradients from these classifiers are combined with the base diffusion score, steering the sampling process toward low-energy regions that also satisfy stability and collision constraints ensuring that the final grasp pairs are dual-arm stable as well as physically valid. The final score is then defined as,
\begin{equation}
\begin{aligned}
\tilde{s}(H, P, t) 
&= s_\alpha(H, P, t) + \nabla_H \log C_\beta^{\text{FC}}(H, P)\\[3pt]
& + 
\begin{cases}
    0, 
    & \text{if } t < t_c, \\[2pt]
    \nabla_H \log \!\left(1 - C_\gamma^{\text{Col}}(H, P)\right),
    & \text{if } t \geq t_c,
\end{cases}
\end{aligned}
\label{eq:guided-score-piecewise}
\end{equation}
where $t_c$ denotes a predefined threshold, after which collision guidance is activated to progressively refine the generated grasps, as refinement is unnecessary while the grasps remain in free space. The corresponding reverse update then follows the same formulation defined in Equation~\ref{eq:expmap-step}.


\begin{figure*}[t]
    \centering
    \vspace{3pt}
    \captionsetup{font=footnotesize}
    \includegraphics[width=0.99\linewidth]{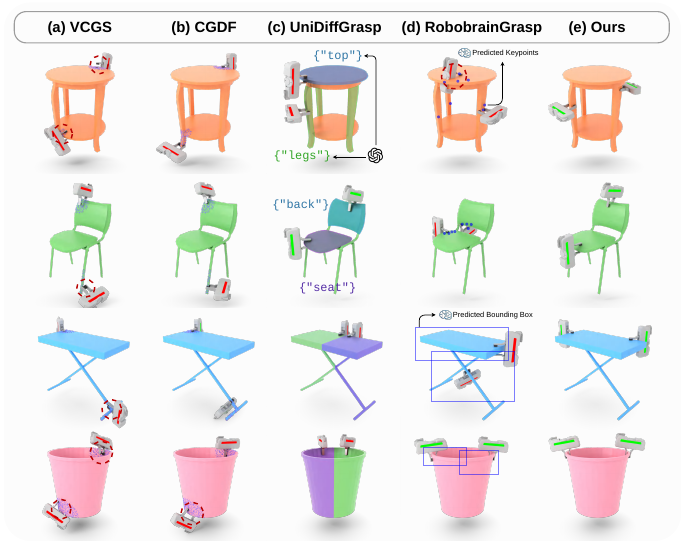}
    \caption{\textbf{Qualitative comparison of dual-arm grasps.} 
    VCGS often fails due to poor grasp generation and its farthest-region heuristic. CGDF shares the latter limitation, for example, on the bucket, one gripper cannot reach the bottom corner. UniDiffGrasp relies on GPT-4V and VLPart for semantic segmentation, but when region labels are ambiguous (e.g., last two rows), it defaults to naive splits, yielding unstable grasps. RoboBrainGrasp predicts keypoints or bounding boxes, yet the keypoints are frequently misaligned (e.g., inside the stool) and the boxes too coarse for precise grasping. In contrast, our method directly generates stable, collision-free dual-arm grasps by reasoning over physical constraints, without heuristics or vague semantic cues. (\textit{\textcolor{darkred}{Red} circles indicate grasps that either collide or fail to contact the object. RoboBrainGrasp-KP and -BB are referred to jointly as RoboBrainGrasp.})}
    \label{fig:qualitative}
\end{figure*}
\subsection{Model Architecture}
\label{sec:model_architecture}
Our framework (Figure \ref{fig:pipeline}) extends diffusion-based grasp generation to the dual-arm setting by introducing a geometry-aware Vision Encoder and a set of specialized Output Heads that jointly enforce feasibility, stability, and collision-free interaction.

\textbf{Feature Encoding. } Inspired from \cite{cgdf}, the input point cloud $P$ is encoded using a VN-PointNet \cite{vnn} to extract $SO(3)$-equivariant per-point features, which are further processed through multi-plane projections and a UNet backbone \cite{unet}. Next, dual-arm grasp poses are transformed by a fixed query point cloud $P_q \in \mathbb{R}^{30\times3}$ to get $P_H$, which represents the local grasp region on the object point cloud. We then retrieve plane features at the projected locations of $P_H$ using bilinear interpolation. These features are then aggregated and passed through the feature encoder $F_\theta$ that conditions on the diffusion step $t$ and jointly predicts (i) a feature vector representation and (ii) the SDF of the query points for geometric supervision. 

\textbf{Multi-Head Output}. The resulting feature vector is then passed through three heads that play distinct yet tightly coupled roles in the generation step:

1. \textit{Energy Head}: It outputs a scalar energy $E_\alpha(H, P, t) \in \mathbb{R}$, where lower energy corresponds to more physically valid grasps. It serves as the backbone of the diffusion process: during training, its gradients teach the model to denoise noisy samples toward low-energy configurations, and during inference, it provides the base generative dynamics.

2. \textit{Force-Closure (FC) Head}: It predicts the probability that a candidate grasp pair achieves force-closure. Importantly, it is trained jointly with the Energy Head, ensuring that the backbone generative process is directly coupled with physically meaningful stability supervision. During inference, its gradients act as guidance signals that bias generation toward dual-arm stable regions of the pose space.

3. \textit{Collision Head}: It predicts the probability of either gripper intersecting the object point cloud. This head is trained after the Vision Encoder, Energy, and FC heads have converged, allowing it to specialize in detecting fine-grained collisions without disturbing the generative objectives. During inference, its gradients are activated after an initial denoising stage, progressively refining candidate grasps by pushing them away from collisions. While our network also predicts SDFs, these are not used for collision refinement since differentiable collision detection is non-trivial and SDF gradients do not guarantee effective grasp refinement.

Together, the three heads form a cooperative architecture: the Energy Head drives diverse and physically valid generation, the FC Head enforces joint grasp stability, and the Collision Head ensures geometric validity. Through this, our framework overcomes the shortcomings of prior methods \cite{unidiff, cgdf} that treat dual-arm grasping as a combination of independent single-arm grasp proposals.

\section{Experiments}
In this section, we describe the dataset, evaluation metrics and the performance of our method compared with different baselines. First, we uniformly sample 1000 points from each object mesh to construct the input point cloud used during inference. For each run, a batch of $B$ dual-arm grasps $\{{H_i \in SE(3) \times SE(3)}\}_{i=1}^B$ are randomly initialized , and the diffusion process is applied for $T = 250$ denoising steps using the formulation defined in Section \ref{sec:methods}. We use the last 50 steps for collision refinement, since by this stage most grasps have converged to feasible regions, with only a small subset intersecting slightly with the object surface. The importance of this refinement stage is further supported by the ablation study in Section \ref{sec:results}. After denoising and refinement, the final set of generated dual-arm grasps are evaluated using the metrics discussed later in this section.

\begin{figure}[!t]
    \centering
    \vspace{5pt}
    \includegraphics[width=\linewidth]{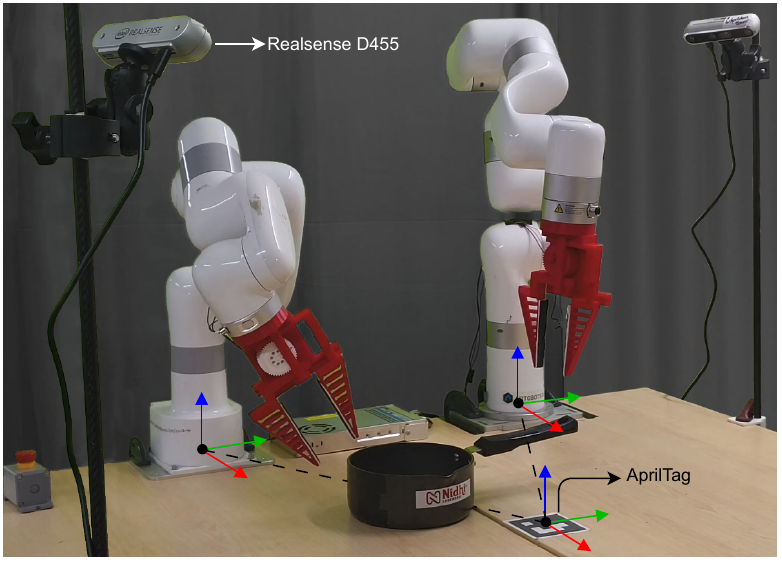}
    \captionsetup{font=footnotesize} 
    \caption{\textbf{Real-world experimental setup.} Dual-arm system with an \textit{XArm7} and \textit{XArm6 Lite}, calibrated using an AprilTag and observed by two RealSense D455 cameras.}
    \label{fig:real_life_setup}
\end{figure}

\textbf{Dataset}: We train and evaluate our model and baselines on the DG16M dataset \cite{dg16m}. The dataset contains 4,143 objects, each consisting approximately 2,000 positive and negative dual-arm grasps validated under improved force-closure evaluation. For our experiments, we adopt a random split in which 400 objects are reserved for testing and the remaining objects are used for training. All reported quantitative results are evaluated on this unseen test split. In addition, we construct a synthetic dataset of colliding and non-colliding dual-arm grasps by sampling grasp poses and checking for collisions with the object. This is used exclusively to train the Collision Head.



\textbf{Baselines.} To assess the performance of our framework, we compare against three categories of baselines for dual-arm grasping: (i) \emph{Farthest-region grasping}, (ii) \emph{VLM-region based grasping}, and (iii) \emph{Affordance-based grasping}.  

\emph{Farthest-region grasping.} \textbf{CGDF} \cite{cgdf} applies a part-guided strategy to generate grasps in the two farthest regions of the point cloud and combines them into dual-arm pairs. Additionally, \textbf{VCGS} \cite{vcgs} uses a variational grasp generation model to generate single-arm grasps in constrained regions and we use it in the same setup as used in \cite{cgdf} for evaluation.

\emph{VLM-based region grasping.} \textbf{UniDiffGrasp} \cite{unidiff} uses GPT-4V \cite{gpt4} and VL-Part \cite{vl_part} to predict two graspable regions on the object and then applies the CGDF part-guided strategy to form dual-arm pairs. To also study the effect of different VLMs, we adapt this baseline by predicting the regions through bounding-box and keypoint prediction using RoboBrain 2.0 7B \cite{robobrain2.0}, a model trained on embodied reasoning tasks. We call these adapted baselines as \textbf{RoboBrainGrasp-BB} and \textbf{RoboBrainGrasp-KP} respectively.  

\emph{Affordance-based grasping.} \textbf{DualAfford} \cite{dualafford} predicts functional object regions for dual-arm interaction and generates the grasp pairs by sequentially conditioning the second gripper’s pose on the first. We did not retrain this baseline on our evaluation dataset due to its complex and object category-specific training pipeline. Instead, we evaluated our framework in a zero-shot manner on the test category objects used in DualAfford’s pickup task.

\textbf{Metrics.} We evaluate the performance of the proposed method and baselines using \textit{Force Closure Evaluation (FCE)}, \textit{Grasp Success Rate (GSR)} and \textit{Grasp Collision Rate (GCR)}. These metrics together capture analytical grasp stability, physical robustness, and geometric feasibility. 

\textit{Force Closure Evaluation (FCE)}: Following the same formulation in DG16M \cite{dg16m}, this analytical metric verifies whether a grasp achieves force closure by testing if the applied contact forces can resist arbitrary external wrenches under friction and gripper force constraints. FCE provides a theoretical guarantee of stability and is the primary analytic measure of our grasp quality.

\textit{Grasp Success Rate (GSR):} This simulation-based metric evaluates the physical robustness of generated grasps in Isaac Gym \cite{isaac_gym}. Each grasp is executed by initializing floating grippers at the predicted grasp pose, closing the fingers, and then enabling gravity. A trial is marked as successful if the object is lifted to a defined height and remains stably grasped. GSR captures whether grasps that satisfy analytic criteria also translate to positive executions under dynamics.

\textit{Grasp Collision Rate (GCR):} This metric measures the percentage of generated grasps whose final pose intersects with the object geometry. A lower collision rate indicates better geometric validity and effectiveness of the collision-guidance mechanism.

\section{Results and Ablation}
\label{sec:results}
\textbf{Qualitative Results.} Figure~\ref{fig:qualitative} illustrates qualitative comparisons across methods. For CGDF and VCGS (\hyperref[fig:qualitative]{Figure~\ref*{fig:qualitative} (a), (b)}), the farthest-region heuristic often selects physically incompatible grasp regions, such as on the bucket where one gripper is forced to an unstable corner. Even when the chosen regions happen to be graspable, like in the case of the chair and stool, the resulting configurations require excessive or unbalanced contact forces. For the VLM-based baselines, as seen in the stool and ironing-table, RoboBrain-based methods (\hyperref[fig:qualitative]{Figure~\ref*{fig:qualitative}(d)}) predict coarse bounding boxes and keypoints that either lie in free space or fail to align with usable regions. Similarly, UniDiffGrasp (\hyperref[fig:qualitative]{Figure~\ref*{fig:qualitative}(c)}) often defaults to arbitrary splits (e.g., the bucket), which result in unstable grasps. In contrast, \coolname (\hyperref[fig:qualitative]{Figure~\ref*{fig:qualitative}(e)}) consistently produces dual-arm grasps that are both stable and collision-free, without relying on heuristics or semantic labels and performs twice as good compared to previous methods in all three metrics. Our classifier-guided diffusion discovers physically consistent grasp regions directly, enabling reliable dual-arm grasp generation across diverse and complex object geometries.

\textbf{Quantitative Results.} Table~\ref{tab:dual_arm_results} summarizes the performance of \coolname compared to other baselines. Farthest-region heuristics (CGDF, VCGS) perform poorly in FCE and GSR because they decouple grasp selection from stability. VCGS further suffers from its reliance on a global shape representation, which explains its very high GCR. VLM-based methods (UniDiffGrasp, RoboBrainGrasp) also show poor performance in all metrics, as their semantically predicted regions are frequently coarse or misplaced due to limited 3D and physical reasoning which leads to grasps generated in the unsuitable object regions. Finally, compared to DualAfford~\cite{dualafford}, which reports results from category-specific models, our single unified model achieves higher GSR in a zero-shot setting (under the same evaluation protocol), indicating stronger generalization to unseen categories. Overall, by coupling diffusion with explicit stability and collision guidance, \coolname overcomes the core limitations of heuristic, semantic, and affordance-based baselines, producing grasps that are jointly stable and generalizable. As shown in Table~\ref{tab:dual_arm_results}, it achieves 60.1\% FCE, 72.5\% GSR, and 15.1\% GCR, roughly twice the stability and success of the baselines, while reducing collisions by more than half.

\begin{table}[!t]
    \centering
    \renewcommand{\arraystretch}{0.95} 
    \setlength{\tabcolsep}{3.5pt} 
    \begingroup
    \normalsize
    \begin{tabular}{lccc}
        \addlinespace[1.5mm]
        \hline
        \textbf{Method}\rule{0pt}{3ex} & \textbf{FCE}(\%)$\uparrow$ & \textbf{GSR}(\%)$\uparrow$ & \textbf{GCR}(\%)$\downarrow$ \\
        \addlinespace[1mm]
        \hline
        \addlinespace[1mm]
        \small DAGDiff (ours) & \small \textbf{60.14} & \small \textbf{72.50} & \small \textbf{15.10} \\
        \small CGDF \cite{cgdf} & \small 35.14 & \small 56.25 & \small 30.55 \\
        \small VCGS \cite{vcgs} & \small 16.85 & \small 23.36 & \small 74.73 \\
        \small UniDiffGrasp \cite{unidiff} & \small 10.10 & \small 31.68 & \small 59.90 \\
        \small RoboBrainGrasp-KP \cite{robobrain2.0} & \small 9.80 & \small 27.85 & \small 66.30 \\
        \small RoboBrainGrasp-BB \cite{robobrain2.0} & \small 7.12 & \small 27.81 & \small 70.26 \\
        \addlinespace[1.2mm]
        \hline
        \addlinespace[1.2mm]
        \small Ours-DA$^\dagger$ & \small \textbf{56.45} & \small \textbf{68.80} & \small \textbf{18.59} \\
        \small Dual-Afford$^{\dagger\dagger}$ \cite{dualafford} & \small -- & \small 54.33 & \small -- \\
        \hline
    \addlinespace[1mm]
    \multicolumn{4}{l}{\scriptsize $^\dagger$Evaluated on Dual-Afford objects in a zero-shot setting.} \\
    \multicolumn{4}{l}{\scriptsize $^{\dagger\dagger}$Values reported directly from the DualAfford paper.}
    \end{tabular}
    \endgroup
    \captionsetup{font=footnotesize} 
    \caption{Comparison of dual-arm grasp generation methods. Higher is better for FCE and GSR; lower is better for GCR.}
    \label{tab:dual_arm_results}
\end{table}

\textbf{Ablations.} The key design choice in our framework is the use of classifier gradients to guide grasp generation, ensuring that the final dual-arm grasps are physically valid and stable. To assess the necessity of this design, we conduct ablation studies to study its contribution, as shown in Table \ref{tab:ablation}.

\textit{(A) Generation without the Force-Closure Head}: In this variant, we remove the Force-Closure head and rely solely on the Energy head’s denoising objective for grasp generation. This leads to a significant drop in both FCE and GSR, indicating that the Energy head alone cannot reliably capture dual-arm stability. Because the FC head is explicitly trained to discriminate between stable and unstable dual-arm grasp pairs, its gradients provide precise and meaningful guidance during generation. This highlights the necessity of explicit force-closure guidance for producing stable dual-arm grasps.

\begin{table}[!t]
    \centering
    \renewcommand{\arraystretch}{0.95} 
    \setlength{\tabcolsep}{4pt} 
    \begingroup
    \normalsize
    \begin{tabular}{lccc}
        \addlinespace[1.5mm]
        \hline
        \textbf{Variant}\rule{0pt}{3ex} & \textbf{FCE}(\%)$\uparrow$ & \textbf{GSR}(\%)$\uparrow$ & \textbf{GCR}(\%)$\downarrow$ \\
        \addlinespace[1mm]
        \hline
        \addlinespace[1mm]
        \small DAGDiff (ours) & \small \textbf{60.14} & \small \textbf{72.50} & \small \textbf{15.10} \\
        \small w/o FC Head & \small 24.94 & \small 30.06 & \small 16.85 \\
        \small w/o Collision Head & \small 50.01 & \small 55.67 & \small 23.50 \\
        \small Post-hoc FC Head & \small 32.37 & \small 46.42 & \small 18.36 \\
        \addlinespace[1.2mm]
        \hline
    \end{tabular}
    \endgroup
    \captionsetup{font=footnotesize}
    \caption{Ablation study with three variants: without force-closure head, without collision head, and with post-hoc FC Head. Metrics reported are FCE, GSR, and GCR.}
    \label{tab:ablation}
\end{table}

\textit{(B) Generation without Collision Head}: Next, we remove the Collision Head and evaluate the grasps without collision refinement, and notice that the collision rate increases from 15.10\% to 23.50\%. This causes the grasps to make incorrect contact with the object surface and this affects FCE and GSR negatively too. The Collision Head provides explicit signals to avoid collisions, which the other heads cannot enforce.


\textit{(C) Post-hoc FC Head training}: To test if the Force-Closure (FC) head could be added post-hoc, we first trained the vision encoder and Energy head, then froze them and trained the FC head separately. This makes FCE and GSR drop substantially, as the encoder could not adapt to the classification objective, leaving the FC head with weak features and poor guidance. This shows that the FC head is not a plug-and-play module but must be jointly trained with the encoder to learn stability-aware representations that effectively guide grasp generation.

    

\begin{table}[!t]
    \centering
    \renewcommand{\arraystretch}{1.0} 
    \setlength{\tabcolsep}{8pt} 
    \begingroup
    \footnotesize
    \begin{tabular}{lccccc}
        \hline
        \textbf{Object}\rule{0pt}{2.5ex} & \textbf{Tray} & \textbf{Bucket} & \textbf{Saucepan} & \textbf{Frypan} & \textbf{Drone} \\
        \addlinespace[1.0mm]
        \hline
        \addlinespace[0.8mm]

        Success & $6/10$ & $8/10$ & $7/10$ & $6/10$ & $5/10$ \\
        \addlinespace[1.0mm]
        \hline
    \end{tabular}
    \endgroup
    \captionsetup{font=footnotesize} 
    \caption{Real-world dual-arm grasp execution results. Each entry shows the number of successful grasps over total attempted grasps for the corresponding object.}
    \label{tab:real_results}
\end{table}

\section{Real Life Experiments}

We validated our framework on a heterogeneous dual-arm setup (Figure \ref{fig:real_life_setup}) with an \textit{XArm7} and an \textit{XArm6 Lite}, using two Intel RealSense D455 cameras for point cloud fusion via ICP. The fused point cloud $P$ was passed to our trained model to generate dual-arm grasps, pruned to retain only kinematically reachable ones. In 10 real-world trials (Table \ref{tab:real_results}), most executions were successful, while failures were mainly from the loss of detail in point cloud reconstruction, which led to incorrect generation of grasps. These results show that our method achieves zero-shot transfer to real sensor data, handling previously unseen objects such as drones and kitchen utensils like saucepans and trays.

\


\vspace{-8pt}
\section{Conclusion}
In this work, we present \coolname, a novel diffusion-based framework for generating stable and collision-free dual-arm grasps directly in the $SE(3) \times SE(3)$ space. By guiding the generative process with force-closure and collision-aware signals, our method outperforms heuristic or region detection-based pipelines and demonstrates reliable zero-shot transfer to previously unseen objects in real-world trials. 
While effective, the framework currently assumes complete, segmented point clouds and does not account for closed chain kinematics. Moreover, its inference speed is limited by the iterative nature of diffusion. As future work, we aim to address these limitations by extending the approach to partial observations, and conducting comprehensive real-world evaluations, paving the way toward more scalable and practical dual-arm manipulation.


\bibliographystyle{IEEEtran}

\bibliography{bibtex}

\end{document}